\documentclass{article} %
\usepackage{iclr2026_conference,times}

\usepackage{amsmath,amsfonts,bm}

\def\eqref#1{equation~\ref{#1}}

\def\1{\bm{1}}

\def\mF{{\bm{F}}}

\def\mL{{\bm{L}}}

\DeclareMathAlphabet{\mathsfit}{\encodingdefault}{\sfdefault}{m}{sl}
\SetMathAlphabet{\mathsfit}{bold}{\encodingdefault}{\sfdefault}{bx}{n}

\def\gF{{\mathcal{F}}}

\def\gL{{\mathcal{L}}}

\def\gP{{\mathcal{P}}}

\usepackage{hyperref}
\usepackage{url}
\usepackage{fontawesome}

\title{HeadsUp! High-Fidelity Portrait Image \\Super-Resolution}

\author{Renjie Li$\bf ^1$, Zihao Zhu$\bf ^1$, Xiaoyu Wang$\bf ^2$, Zhengzhong Tu$\bf ^1$\thanks{tzz@tamu.edu} \\
$^1$Texas A\&M University, $^2$Topaz Labs\\
}

\usepackage{hyperref}       %
\usepackage{url}            %
\usepackage{booktabs}       %
\usepackage{amsfonts}       %
\usepackage{nicefrac}  
\usepackage{colortbl}
\usepackage{microtype}      %
\usepackage{xcolor}         %

\usepackage{wrapfig}
\usepackage{amssymb}%
\usepackage{pifont}%
\usepackage{multirow}
\usepackage{graphicx}

\usepackage{multirow}
\usepackage{amsmath}
\usepackage{enumitem}
\usepackage{xspace}

\newcommand{\ie}{\emph{i.e.}}
\newcommand{\Mat}{\boldsymbol}

\newcommand{\real}{\mathbb{R}}

\newcommand{\ours}{HeadsUp}
\newcommand{\ourdata}{PortraitSR-4K}

\newcommand{\bfours}{\textbf{HeadsUp}}
\newcommand{\bfourdata}{\textbf{PortraitSR-4K}}

\usepackage[capitalize]{cleveref}
\crefname{section}{Sec.}{Secs.}
\Crefname{section}{Section}{Sections}
\Crefname{table}{Table}{Tables}
\crefname{table}{Tab.}{Tabs.}

\definecolor{maroon}{cmyk}{0,0.87,0.68,0.32}
\definecolor{myyellow}{RGB}{218, 160, 109}
\definecolor{brickred}{rgb}{0.8, 0.25, 0.33}
\definecolor{brandeisblue}{rgb}{0.0, 0.44, 1.0}
\definecolor{applegreen}{rgb}{0.55, 0.71, 0.0}
\definecolor{aogreen}{rgb}{0.0, 0.5, 0.0}
\definecolor{mygray}{gray}{0.94}

\definecolor{gdmb}{RGB}{47, 114, 173}  %
\definecolor{gdmr}{RGB}{199, 100,  38}
\definecolor{gdmg}{RGB}{70, 155, 118}
\definecolor{gdmm}{RGB}{193, 126, 165}
\definecolor{gdmy}{RGB}{239, 227,  98}
\definecolor{gdmc}{RGB}{110, 179, 228}
\definecolor{gdmk}{RGB}{20, 20, 20}
\definecolor{turquoise}{cmyk}{0.65,0,0.1,0.3}
\definecolor{purple}{rgb}{0.65,0,0.65}
\definecolor{dark_green}{rgb}{0, 0.5, 0}
\definecolor{orange}{rgb}{0.8, 0.6, 0.2}
\definecolor{red}{rgb}{0.8, 0.2, 0.2}
\definecolor{darkred}{rgb}{0.6, 0.1, 0.05}
\definecolor{blueish}{rgb}{0.0, 0.3, .6}
\definecolor{light_gray}{rgb}{0.7, 0.7, .7}
\definecolor{pink}{rgb}{1, 0, 1}
\definecolor{greyblue}{rgb}{0.25, 0.25, 1}
\definecolor{orgred}{rgb}{1.0, 0, 0}

\iclrfinalcopy %
\begin{document}

\maketitle

\begin{figure*}[h]
    \centering
    \vspace{-7mm}
    \includegraphics[width=0.98\textwidth]{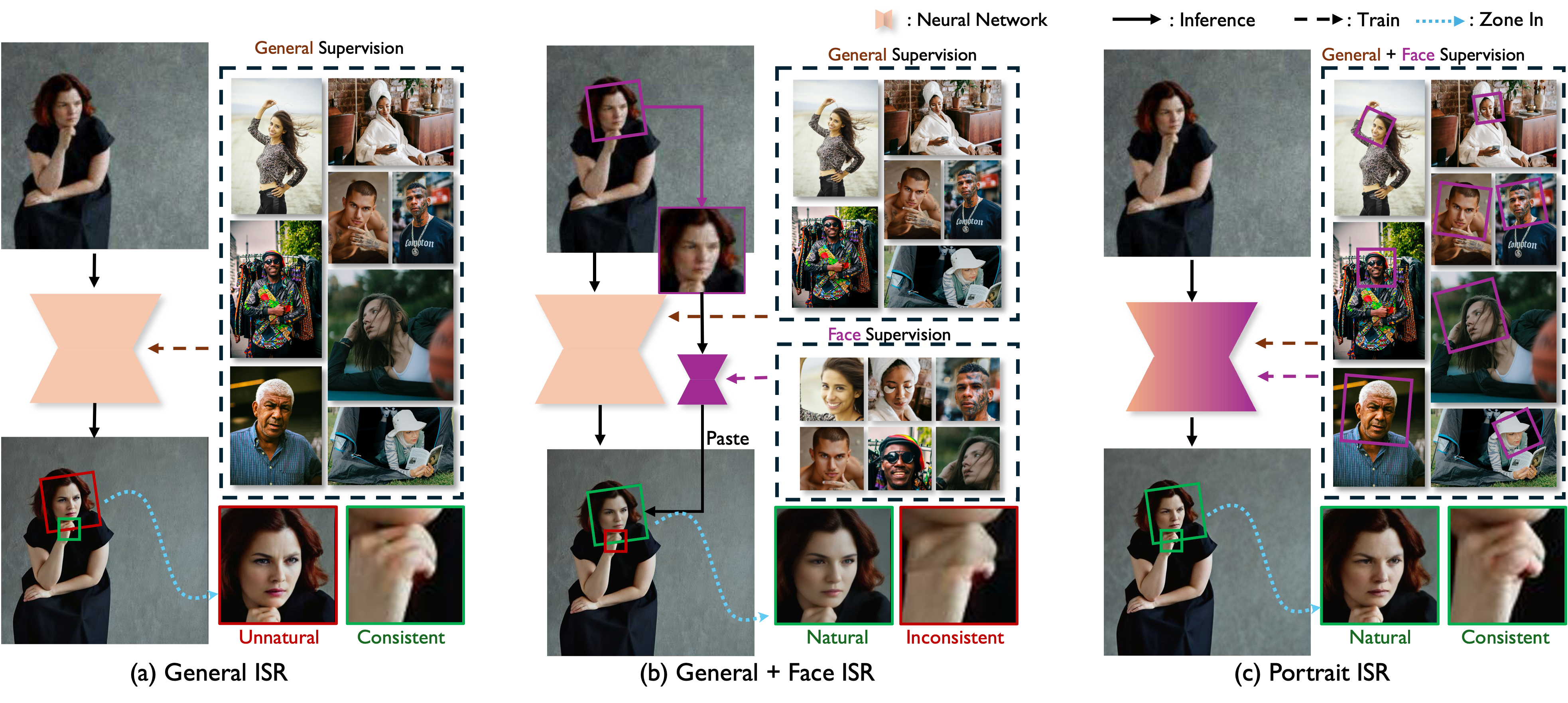}
    \vspace{-1em}
    \caption{Different approaches to solve the portrait image super resolution (ISR) task:
    \textbf{(a)} General ISR models like~\citep{wu2024one} may produce unnatural faces when applied to portrait photos due to the lack of face-specific supervision; \textbf{(b)} While introducing an extra face ISR expert~\citep{zhou2022towards} can generate a more natural face, the \emph{blending} procedure will introduce inconsistent boundaries. \textbf{(c)} Our portrait ISR approach, HeadsUp, can generate a natural portrait photo without introducing inconsistent boundaries around faces using an all-in-one face-aware restoration model.
    }
    \label{fig:teaser}
\end{figure*}

\begin{abstract}
Portrait pictures, which typically feature both human subjects and natural backgrounds, are one of the most prevalent forms of photography on social media. 
Existing image super-resolution (ISR) techniques generally focus either on generic real-world images or strictly aligned facial images (\ie, face super-resolution).
In practice, separate models are blended to handle portrait photos: the face specialist model handles the face region, and the general model processes the rest. 
However, these blending approaches inevitably introduce blending or boundary artifacts around the facial regions due to different model training recipes, while human perception is particularly sensitive to facial fidelity. 
To overcome these limitations, we study the portrait image supersolution (PortraitISR) problem, and propose \bfours, a single-step diffusion model that is capable of seamlessly restoring and upscaling portrait images in an end-to-end manner. 
Specifically, we build our model on top of a single-step diffusion model and develop a face supervision mechanism to guide the model in focusing on the facial region. 
We then integrate a reference-based mechanism to help with identity restoration, reducing face ambiguity in low-quality face restoration.
Additionally, we have built a high-quality 4K portrait image ISR dataset dubbed \bfourdata, to support model training and benchmarking for portrait images. 
Extensive experiments show that \ours\xspace achieves state-of-the-art performance on the PortraitISR task while maintaining comparable or higher performance on both general image and aligned face datasets. 

\end{abstract}
\section{Introduction}
\label{sec:intro}
\vspace{-4mm}

Image super-resolution (ISR) is an essential computer vision task that aims to recover high-quality, high-resolution images from degraded low-quality counterparts.
Significant advancements in ISR have been driven by collecting high-quality image datasets~\citep{cai2019toward,wei2020component}, realistic degradation simulation via the combination of pre-set degradations~\citep{wang2021real, zhang2021designing}, learning from real-world degradation distributions~\citep{wang2021unsupervised,fritsche2019frequency,yuan2018unsupervised}, and leveraging priors from generative foundation models~\citep{wu2024seesr, wang2024exploiting, sun2024improving, wu2024one}.
Despite general success across various image domains, however, existing models often exhibit notably inferior performance when applied to in-the-wild portrait pictures that contain human faces, an area where human perception is especially sensitive to errors in detail and fidelity. 
Portrait images account for a large portion of online photography. Failing to handle both face and background at the same time will cause inconsistency.

To better handle facial images, a variety of methods train \emph{specialist face ISR} models on aligned facial images by leveraging face geometric prior~\citep{yu2018super, shen2018deep, chen2018fsrnet, kim2019progressive}, reference signals~\citep{zhang2024instantrestore, li2020blind, li2020enhanced, li2018learning, chong2025copy}, generative prior~\citep{wang2021towards, yang2025diffusion, wang2024osdface}, and quantized codebooks~\citep{wang2024osdface, zhou2022towards}.
These approaches have substantially improved the reconstruction quality of facial images with better detail and perceptual quality.
However, they are restricted to aligned facial data, limiting their utility for generic portrait photography.
A practical solution to implement portrait ISR is to employ a hybrid fusion strategy~\citep{zhou2022towards, wang2021real}---using a generalist ISR for backgrounds and a specialist face ISR model for facial regions in a crop-project-restore-blend paradigm. 
Unfortunately, this segmented processing approach frequently results in visible boundary artifacts and inconsistencies between facial regions and the surrounding background, significantly degrading the overall perceptual quality of enhanced portraits.

Therefore, we investigate the Portrait Image Super-Resolution problem \textbf{PortraitISR} for short, and propose an \textbf{end-to-end} portrait image super-resolution approach that \ding{182} achieves seamless portrait ISR without any boundary effects, and \ding{183} maintains high-quality background and high-fidelity face as the blending-based methods. 
A na\"ive baseline could be a generalist ISR model trained on portrait data. 
However, achieving such an integrated model is non-trivial due to several identified challenges: \textbf{\underline{Firstly}}, while face is usually the most sensitive part with insufficient details and low-fidelity, it only occupies a small portion of the whole image in most circumstances---\textbf{the small 20\% region gets the big 80\% importance}. Diverse scales, positions, and orientations of faces present in casual captures further worsen this issue. 
\textbf{\underline{Secondly}}, extremely low-quality inputs introduce substantial \textbf{ambiguity}, making precise facial detail reconstruction a particularly challenging ill-posed problem.
\textbf{\underline{Finally}}, while abundant data is available for general ISR~\citep{li2023lsdir, agustsson2017ntire, wang2018recovering} as well as the aligned facial ISR~\citep{liu2015faceattributes, karras2019style} individually, a \textbf{dedicated, high-quality dataset} with diverse in-the-wild faces for PortraitISR tasks remains absent.

In response to the above challenges, we propose \bfours, an end-to-end framework for high-fidelity, face-aware PortraitISR using a single model.
Specifically, we first propose a face-aware region loss that emphasizes both face perceptual quality and face identity. 
Additionally, we design an adaptive face identity module that allows information flow from a reference face image as a promptable identity guidance. 
Finally, we construct a large-scale, high-resolution benchmark, \bfourdata, that contains 30k high-quality 4K portrait images, curated and filtered from web-scale data. 
Experimental evaluations demonstrate that our proposed approach achieves state-of-the-art results in portrait ISR, surpassing existing methods in terms of perceptual quality and fidelity, while maintaining competitive performance on general ISR benchmarks. In summary, our contributions include:
\begin{itemize}[nosep, topsep=0pt, leftmargin=4mm]
    \item We study the \textbf{PortraitISR} problem and introduce \bfours, a novel end-to-end framework specifically designed for seamless portrait image super-resolution without any need for post-processing blending, producing high-quality outputs without any boundary artifacts.

\item We propose a face-aware region loss and a reference-guided adaptive face identity mechanism to improve facial restoration quality significantly, which better trains our proposed diffusion model.

\item We build \bfourdata, the first-of-its-kind, carefully curated, high-resolution (4K) portrait ISR dataset containing 30k images, facilitating future research in portrait ISR tasks.

\item We have established a benchmark on our proposed \bfourdata, where comprehensive experimental results demonstrate superior performance of HeadsUp over existing ISR and face-specific methods. We have also conducted ablation studies to show the design components of our model.
\end{itemize}

\section{Methodology}

\begin{figure}
    \centering
     \includegraphics[width=0.99\linewidth]{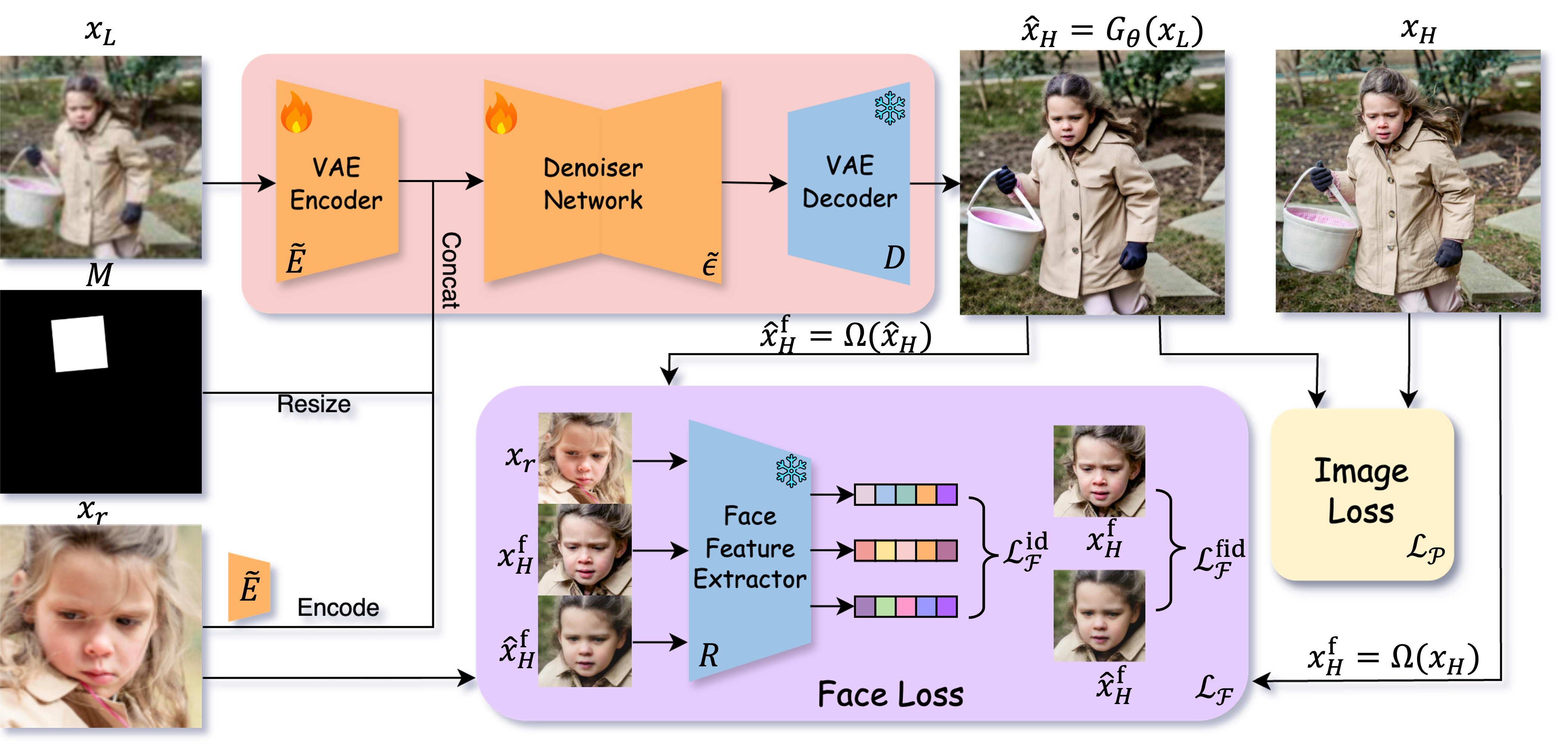}
     \vspace{-3mm}
    \caption{\textbf{Pipeline of \bfours.} Starting from a pre-trained latent diffusion model, we add a LoRA adapter to the VAE encoder and denoising network. \ours~takes as input an LQ image and an optional reference, and denoises for only one step to produce an HQ image. In the training stage, we employ face-specific losses to improve facial restoration quality.}
    \label{fig:pipeline}
    
\end{figure}

\subsection{Preliminary}

To make the representation compact, latent diffusion models (LDMs) represent images in a low-resolution latent space. LDM consists of two procedures. The forward procedure gradually introduces Gaussian noise on the latent codes until the noise-added latent codes are subject to a Gaussian distribution. The forward procedure is denoted as $q(x_t|x_{t-1})=\mathcal{N}(x_t,\sqrt{1-\beta_t}x_{t-1}, \beta_t I)$, where $x_t$ and $x_{t-1}$ are the latent code at step $t$ and $t-1$. To recover the clean latent code $x_0$ from the random noise $x_T$, diffusion models progressively remove noise from $x_T$, described as $p_\theta(x_{t-1}|x_t)=\mathcal{N}(\mu_\theta(x_t, t), \Sigma_\theta(x_t,t))$, where $\mu_\theta$ and $\Sigma_\theta$ are learned denoising functions. In practice, we usually learn a neural network $\epsilon_\theta$ to predict the noise in $x_t$. With DDIM we can jump to any diffusion step $s$ from step $t$ by using $x_s=\alpha_sx_0'+\beta_s\epsilon_\theta(x_t,t)$, $x_0'$ is estimated via $x_0'=\frac{x_t-\beta_t\epsilon_\theta(x_t,t)}{\alpha_t}$. However, one-step diffusion super-resolution models (e.g., OSEDiff~\citep{wu2024one}), in contrast to generative models, often start from a latent code encoded from LQ images rather than a randomly initialized noise, thus they can usually use fewer denoising steps by directly estimating $x_0$ from $x_t$ using DDIM.

\subsection{Portrait Image Super-resolution (PortraitISR)}

\label{sec:pisr}
We follow OSEDiff~\citep{wu2024one} to formulize the general image super-resolution task as $\hat{\Mat x}_H=\arg\min_{\Mat x_H}\gL_{\rm{data}}(\Phi(\Mat x_H), x_L) +  \lambda\gL_{\rm{reg}(\Mat x_H)}$, where $\hat{\Mat x}_H$ is the predicted high-quality (HQ) restored image, $\Mat x_L$ the input low-quality (LQ) image, $\Phi$ the degradation function, $\gL_{\rm{data}}$ the pair-wise supervise term, and $\gL_{\rm{reg}}$ the regularization term. For portrait images, we separately model the entire portrait image and the face region. 
We denote $\gP$ the portrait image set and $\gF$ the set of aligned face images. For each portrait image $\Mat x_p\in\gP$, we can extract its face image and align it with a standard template using an affine transformation, denoted as $\Mat x_f = \Omega(\Mat x_p) \in \gF$, where $\Omega$ is the projection function to transform portrait images to aligned face images. We also introduce a reference facial image $\Mat x_r\in\gF$ to better guide the model. Thus we introduce the task of \underline{portrait image super-resolution} (PortraitISR): 
given input LQ image $\Mat x_L\in\gP$ and an optional reference face image $\Mat x_r \in \gF$ or $\Mat x_r = 0$, estimate an HQ image $\hat{\Mat x}_H$ that (1) the enhanced HQ image should be consistent with the input LQ image; (2) the face in the enhanced image should be consistent with both the LQ face and reference face; (3) the entire HQ image should follow the natural image prior. We modify the portrait ISR formula from the general one as follows:
$$\hat{\Mat x}_H = \operatornamewithlimits{\rm{argmin}}_{\Mat x_H\in\gP}~\lambda_\gP \gL_{\gP}(\Phi_{\gP}(\Mat x_H), \Mat x_L) + \lambda_\gF \gL_\gF(\Phi_{\gF}(\Omega(\Mat x_H)), \Omega(\Mat x_L), \Mat x_r) + \lambda_{\rm{reg}}\gL_{\rm{reg}}(\Mat x_H),$$
where $\gL_{\gP}$ and $\gL_{\gF}$ respectively measures the fidelity on the entire portrait image and the face region; $\gL_{\rm{reg}}$ represents the regularization term the as the general ISR task; $\lambda_\gP$, $\lambda_\gF$ and $\lambda_{\rm{reg}}$ is the coefficients to balance the different terms.

Similar to the general ISR task~\citep{wu2024one}, we learn a neural network model $G_\theta$ to recover the HQ image. The training data is formulized as triplets $\mathcal{S}=\{(\Mat x_L, \Mat x_H, \Mat x_r)|\Mat x_L,\Mat x_H \in \gP, \Mat x_r\in \gF ~\rm{or}~ \Mat x_r=0\}$. The learning problem is described as:
\begin{align}
 \theta^*&=\operatornamewithlimits{argmin}_\theta\mathbb{E}_{(\Mat x_L, \Mat x_H, \Mat x_r)\sim\mathcal{S}}\left[\gL(\Mat x_H, \Mat x_L, \Mat x_r, G_\theta)\right]\\
 \gL &= \lambda_\gP \gL_\gP(\Mat x_H, G_\theta(\Mat x_L))+\lambda_\gF \gL_\gF(\Omega(\Mat x_H), \Omega(G_\theta(\Mat x_L)), \Mat x_r) + \lambda_{\rm{reg}}\gL_{\rm{reg}}(G_\theta(\Mat x_L)),
\label{eq:objective}
\end{align}
where $\gL_\gP, \gL_\gF$ and $\gL_{reg}$ are the loss or regularize terms. In the following section, we will introduce the detailed design of the loss functions and the model.

\subsection{Face Aware Supervision}
\label{sec:face_supervision}
As mentioned above, simply training an ISR model on portrait data can face the problem of insufficient supervision on face images and ambiguity for face identities. To tackle these problems, we carefully design the face objectives $\gL_\gF$ in Eq.~\ref{eq:objective} while the other terms are derived from~\citep{wu2024one}, which includes MSE, LPIPS, and VSD loss. Specifically, the face loss $\gL_\gF$ consists of three parts, the face fidelity loss $\gL_\gF^{\rm{fid}}$, the reference involved face identity loss $\gL_\gF^{\rm{id}}$, and the face adversarial loss $\gL_\gF^{\rm{adv}}$, denoted as:
$$\gL_\gF(\Mat x_{H}^{\rm{f}},\hat{\Mat x}_{H}^{\rm{f}}, \Mat x_r)=\lambda^{\rm{fid}}\gL_\gF^{\rm{fid}}(\Mat x_{H}^{\rm{f}}, \hat{\Mat x}_{H}^{\rm{f}})+\lambda^{\rm{id}}\gL_\gF^{\rm{id}}(\Mat x_{H}^{\rm{f}}, \hat{\Mat x}_{H}^{\rm{f}}, \Mat x_r)
+\lambda^{\rm{adv}}\gL_\gF(\Mat x_{H}^{\rm{f}}, \hat{\Mat x}_{H}^{\rm{f}}, \Mat x_r),
$$
where $\Mat x_{H}^{\rm{f}}=\Omega(\Mat x_H))$ the cropped and aligned face region in the HQ image, $\hat{\Mat x}_{H}^{\rm{f}}=\Omega(G_\theta(\Mat x_L)$ the face image of the model prediction, and $\Mat x_r$ the reference face image.
\paragraph{Face Fidelity Loss.} To learn a model that can restore portrait images with a more fine-grained face, we employed a face fidelity loss that is specifically applied on the aligned face region, denoted as 
\begin{equation}
\gL_\gF^{fid}(\Mat x_{H}^{\rm{f}}, \hat{\Mat x}_{H}^{\rm{f}}) = ||\Mat x_{H}^{\rm{f}} - \hat{\Mat x}_{H}^{\rm{f}}||_2 + \lambda_{\rm{LPIPS}}\gL_{\rm{LPIPS}}(\Mat x_{H}^{\rm{f}}, \hat{\Mat x}_{H}^{\rm{f}}).
\end{equation}
\paragraph{Face Identity Loss.} To guide the model to preserve the identity of the face region, we developed an identity loss. We further introduce a reference face image into the system to deal with the ambiguity problem. To model face identity, inspired by~\citep{wang2024edit}, we employ an off-the-shelf face recognition model $R:\gF\rightarrow\real^d$ as a feature extractor, then we build a pair-wise identity criterion based on the cosine similarity of their recognition features as follows:
\begin{equation}
\varphi(\Mat x, \Mat y)=\frac{\langle R(\Mat x), R(\Mat y)\rangle}{||R(\Mat x)||_2 \cdot||R(\Mat y)||_2} \in \left[-1, 1\right]
\label{eq:identity_sim}
\end{equation}
where $\Mat x$, $\Mat y$ are two aligned facial images, $\langle\cdot\rangle$ the inner production, and $||\cdot||_2$ the L2-norm of vector.
Then we construct the identity loss considering the HQ face $\Mat x_H^{\rm{f}}\in\gF$, predicted face $\hat{\Mat x}_H^{\rm{f}}\in \gF$ and the optional reference image $\Mat x_r\in\gF$ or $\Mat x_r=0$.
The intuition is that the predicted face should always be similar to the HQ face (the GT term). If there is a reference face, the identity loss should encourage the model to take information from the reference face by explicitly taking the identity similarity of the predicted and reference faces into account (the reference term). Further, we weight the terms by the similarity between the HQ face and the reference face. The identity loss is formalized as follows:
$$\gL_\gF^{\rm{id}}(\Mat x_{H}^{\rm{f}}, \hat{\Mat x}_{H}^{\rm{f}},\Mat x_r)=-\log(\frac{\varphi(\hat{\Mat x}_{H}^{\rm{f}}, \Mat x_{H}^{\rm{f}})+1}{2}) -\log(\frac{\varphi(\hat{\Mat x}_{H}^{\rm{f}}, \Mat x_r)+1}{2})\varphi(\Mat x_H^{\rm{f}}, \Mat x_r).$$
Note that if there is no reference image, we define its similarity with any face to zero, i.e., $\varphi(\Mat x, 0)=0, \forall \Mat x \in \gF$. In this case, the identity loss will have the GT term only.

\subsection{One-step PortraitISR Framework}
\paragraph{Overview.}

As mentioned in~\ref{sec:pisr}, 
the portrait ISR model $G_\theta$ takes as input an LQ image $\Mat x_L \in \gP$ and an optional reference image $\Mat x_r\in\gF$ or $\Mat x_r=0$. If the reference image is provided, we additionally introduce a binary mask $M\in\{0,1\}^{h\times w}$ to specify the face location in the LQ image, where $h$ and $w$ are the height and width of the LQ image. The model then predicts the HQ portrait image $\hat{\Mat x}_H$ from the given conditions.
Following~\citep{wu2024one}, we build $G_\theta$ as a one-step diffusion model. As it also requires a text prompt as input, we employ a text extractor $T:P\rightarrow\mathcal{T}$ to estimate the corresponding text prompt from the LQ image, where $\mathcal{T}$ is the text set. The portrait ISR procedure is formalized as $\hat{\Mat x}_H=G_\theta(\Mat x_L, \Mat x_r, M, T(\Mat x_L))$.

\paragraph{Architecture.} As shown in Fig.~\ref{fig:pipeline}, we start from a pre-trained latent diffusion model $\Psi=(E,\epsilon, D)$, where $E$, $\epsilon$, $D$ represent the VAE encoder, the denoiser, and the VAE decoder, respectively. We fixed the decoder and apply a LoRA~\citep{hu2022lora} adaption to he encoder and denoiser, denoted as $G_\theta = (\Tilde{E}_{\theta1}, \Tilde{\epsilon}_{\theta2}, D)$, where $\Tilde{M}_\theta$ means adding a LoRA adapter parameterized by $\theta$ to the module $M$. We omit the subscript $\theta$ in the following. Further, to enable the denoiser with the reference latent and mask as input, we extend its first convolution layer with additional zero-initialized filter channels.
The inference procedure is described as follows: (1) the LQ image and reference image are encoded into latent space $\Mat z_L=\Tilde{E}(\Mat x_L)$, $\Mat z_r = \Tilde{E}(\Mat x_r)$. (2) the latents and the resized face mask are concatenated and the denoised for one step $\hat{\Mat z}_H=\frac{\Mat z_L - \beta\Tilde{\epsilon}(\Mat z_L,\Mat z_r, M^{\rm{resize}})}{\alpha}$, where $\alpha$ and $\beta$ are the diffusion scalar and $M^{\rm{resize}}$ is the face mask resized to latent resolution. (3) The denoised latent are decoded into HQ portrait image $\hat{\Mat x}_H=D(\hat{\Mat z}_H)$.

\section{\ourdata~Dataset}
\label{sec:data_collection}
While large-scale datasets~\citep{li2023lsdir, agustsson2017ntire, wang2018recovering, liu2015faceattributes, karras2019style} have significantly facilitated the ISR and FSR field, the lack of a high-quality portrait image dataset limits the development of PortraitISR approaches. In this work, we propose \bfourdata, a large-scale portrait dataset which consists of 30k high-quality 4K portrait images from the internet.
\paragraph{Image Collection}
We collected the raw videos from existing datasets, including Laion2B~\citep{schuhmann2022laion}, Photo Concept Bucket\footnote{https://huggingface.co/datasets/bghira/photo-concept-bucket}, and PD12M~\citep{meyer2024public}. We selected the images with at least 4K resolution as our raw image candidates, denoted as $\gP$.

\paragraph{Portrait Data Construction.} 
After collecting the raw image set, we employ a face detector $\phi$ to detect the faces in each of the images. We construct the portrait set $ \Mat P \subset \gP$ by removing the images that include no face or the face is too small. 
We then cropped and aligned the detected faces via affine transformation to form the face image set $\Mat F$. We construct the reference pairs $\Mat R$ by employing the identity criteria in Eq.~\ref{eq:identity_sim} on each face pairs and collecting the pairs whose similarity is above certain threshold $\gamma$, i.e. $\Mat{R} = \{(\Mat x, \Mat y)|\Mat x\in \Mat F, \Mat y \in \Mat F, \varphi(\Mat x, \Mat y) > \gamma\}$. The training pairs $\mathcal{S}$ is constructed as $\mathcal{S}=\{(\Phi(\Mat x), \Mat x, \Mat y) |\Mat x\in \Mat F, (\Omega(\Mat x), \Mat y)\in\Mat{R}\}$, where the terms in the triplet represents the LQ portrait, the HQ portrait, and the reference face image.

\paragraph{Dataset Splitting}
We collected 30k portrait data. The training set (\ourdata-Tr) consists of 27k images, and the testing set (\ourdata-Te) consists of 3k images. We construct pairs within each subset. For the training set, all valid pairs are recorded, totaling  163k training pairs. For the testing set, each face will have at most one reference face; if multiple similar faces are detected, we simply select the most similar one. Finally, we reserved 190 portrait-reference testing pairs.

\begin{figure}[t]
    \centering
\includegraphics[width=0.95\linewidth]{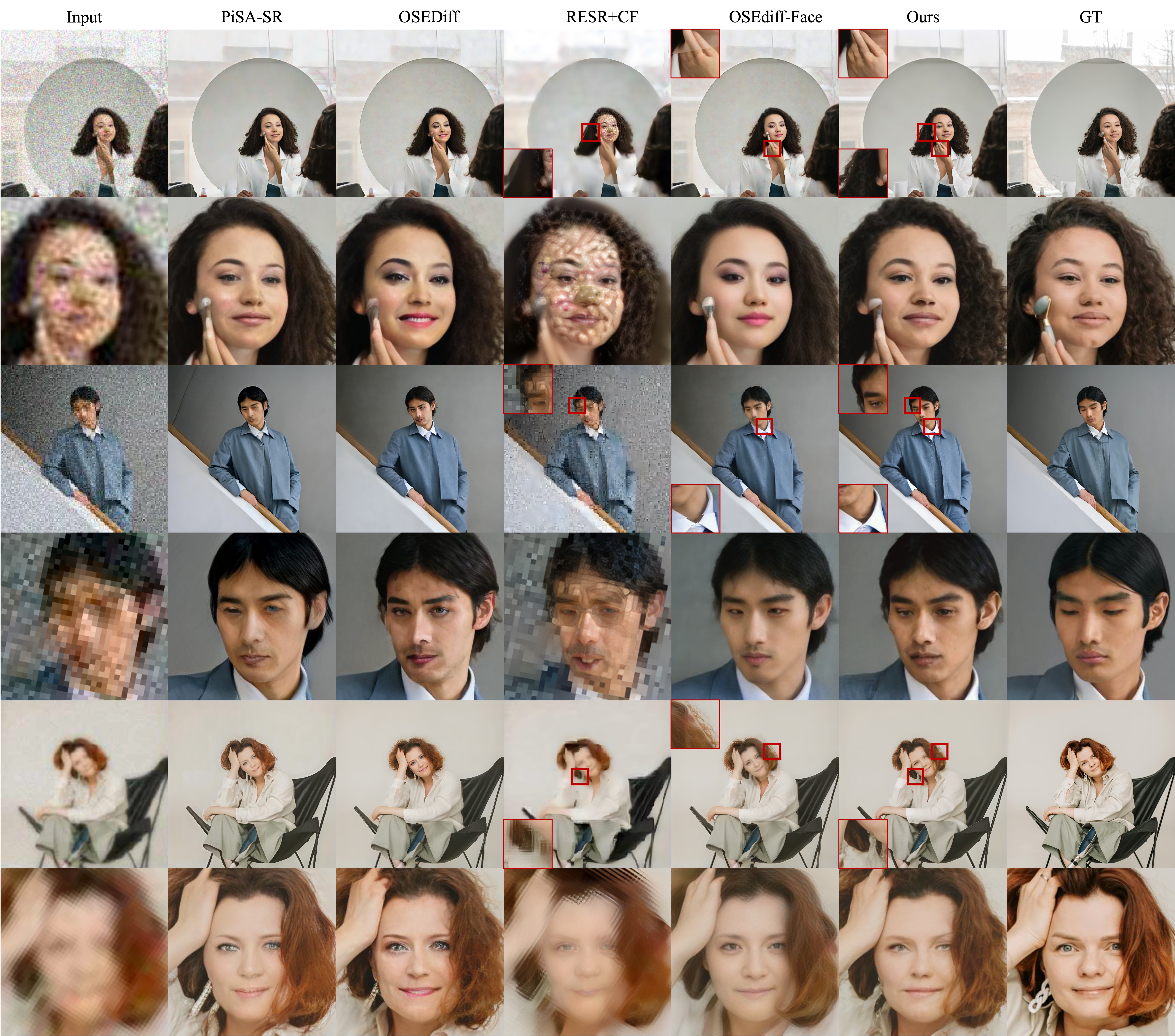}
     \vspace{-3mm}
    \caption{\textbf{Qualitative Results.} While general ISR approaches can achieve good overall quality, they can not produce high-fidelity faces. The blending approaches produce better face fidelity, but suffer from the border effect that causes inconsistency between the face and other regions. }
    \label{fig:compare}
\end{figure}

\section{Experiments}
\subsection{Experimental Settings}
\label{sec:ex_settings}

\paragraph{Datasets.} \underline{Training:} To maintain a fair performance on general ISR tasks, we train our model on a mixture of \ourdata-Tr, LSDIR~\citep{li2023lsdir}, and DIV2K~\citep{agustsson2017ntire}. The training resolution is at $1024 \times 1024$, and the degradation pipeline is derived from Real-ESRGAN~\citep{wang2021real}. \underline{Testing:} For portrait ISR, we test our model on our \ourdata-Te, with the input resolution at $256\times 256$. For general ISR, we follow~\citep{wu2024one} to evluate the models, consisting of images from DIV2K-VAL~\citep{agustsson2017ntire}, RealSR~\citep{cai2019toward}, and DRealSR~\citep{wei2020component}, whose LQ and HQ image is at $128\times 128$ and $512\times512$ resolution, respectively. For FSR, following ~\citep{tsai2024dual}, we choose celeba-Test~\citep{liu2015faceattributes} as the testing set.
\paragraph{Baselines.} 
\begin{wraptable}[14]{c}{0.55\textwidth}
\vspace{-7mm}
\caption{\textbf{Comparison on Face ISR.} We compare \ours~with specialist face restoration models. We achieve the best performance on many metrics, with the rest of the metrics comparable to other methods. The top two performances are labeled in {\color{red}red} and {\color{greyblue}blue}.}
\label{tab:fsr}
\vspace{3pt}
\resizebox{0.55\textwidth}{!}{
\begin{tabular}{ccccccc}
\toprule
Method        & PSNR$\uparrow$   & SSIM$\uparrow$  & LPIPS$\downarrow$ & FID$\downarrow$    & NIQE$\downarrow$  & IDA$\downarrow$   \\
\midrule
PSFRGAN       & 20.303 & 0.536 & 0.450 & 66.367 & \textbf{\color{greyblue}3.811} & 1.260 \\
GFP-GAN       & 19.574 & 0.522 & 0.453 & \textbf{\color{greyblue}46.130} & 4.061 & 1.268 \\
GPEN          & 20.545 & 0.552 & 0.425 & 55.308 & 3.913 & 1.141 \\
RestoreFormer & 20.146 & 0.494 & 0.467 & 54.395 & 4.013 & 1.231 \\
CodeFormer    & \textbf{\color{greyblue}21.449} & 0.575 & \textbf{\color{greyblue}0.365} & 62.021 & 4.570 & \textbf{\color{red}1.049} \\
VQFR          & 19.484 & 0.472 & 0.456 & 54.010 & \textbf{\color{red}3.328} & 1.237 \\
DR2           & 20.327 & \textbf{\color{greyblue}0.595} & 0.409 & 63.675 & 5.104 & 1.215 \\
DAEFER        & 19.919 & 0.553 & 0.388 & 52.056 & 4.477 & \textbf{\color{greyblue}1.071}\\
\cellcolor{mygray}Ours & \cellcolor{mygray}\textbf{\color{red}22.924}  & \cellcolor{mygray}\textbf{\color{red}0.684}  & \cellcolor{mygray}\textbf{\color{red}0.345}  &  \cellcolor{mygray}\textbf{\color{red}29.226} & \cellcolor{mygray}4.264 & \cellcolor{mygray}1.103 \\
\bottomrule

\end{tabular}
}
\end{wraptable}

\underline{PortraitISR:} To the best of our knowledge, there is no PortraitISR specialist model, thus we employed two kinds of baselines: (1) The General ISR (GISR) approaches, including OSEDiff~\citep{wu2024one}, 
and PiSA-SR~\citep{sun2024pixel}. (2) The practical blending approaches, which are the combination of an ISR and an FSR model. Specifically, we choose Real-ESRGAN~\citep{wang2021real} + CodeFormer~\citep{wu2024seesr} and OSEDiff~\citep{wu2024one} + OSEDiff\_Face~\citep{wu2024one} as the baselines. \underline{GISR:} Following PiSA-SR~\citep{sun2024pixel}, we choose state-of-the-art diffusion-based methods~\citep{yue2023resshift,wang2024exploiting,lin2024diffbir,yang2024pixel,wang2024sinsr,wu2024seesr, wu2024one,sun2024pixel} as baselines.  \underline{FSR}: We align our baseline models with recent DEAFR~\citep{tsai2024dual}. The baselines are set to PSFRGAN~\citep{chen2021progressive}, GFP-GAN~\citep{wang2021towards}, GPEN~\citep{yang2021gan}, RestoreFormer~\citep{wang2022restoreformer}, CodeFormer~\citep{zhou2022towards}, VQFR~\citep{gu2022vqfr}, DR2~\citep{wang2023dr2}, DAEFR~\citep{tsai2024dual}.

\paragraph{Metrics.} Following~\citep{wu2024one}, we employ PSNR, SSIM~\citep{wang2004image}, LPIPS~\citep{zhang2018unreasonable}, FID~\citep{heusel2017gans}, and NIQE~\citep{zhang2015feature} as evaluation metrics for all tasks. For the FISR task, we add the IDA~\citep{deng2019arcface} metric to align with in~\citep{tsai2024dual}. For the GISR task, we further use MUSIQ~\citep{ke2021musiq}, CLIPIQA~\citep{wang2023exploring}, DISTS~\citep{ding2020image} and MANIQA~\citep{yang2022maniqa} as the evaluating metrics, which is consistent with~\citep{wu2024one}. For PortraitISR, we employed MUSIQ~\citep{ke2021musiq}, DISTS~\citep{ding2020image} and MANIQA~\citep{yang2022maniqa}. 
We employ the similarity score as defined in Eq.~\ref{eq:identity_sim} to evaluate the identity similarity. However, the cosine distance-based metric may be noisy; we further conduct user studies to measure the identity similarity. 
We conducted extra user studies to evaluate the identity similarity and the naturalness of the faces produced by different models. In the user study, the participants are given several candidate face images, and they are asked to select one image that best meets a certain criterion. We report the ``win rate (WR)'' as the user study metric. The win rate of one method is the frequency it is selected as the best image, out of all the selections. We denote the face naturalness metric as $\rm{WR}_N$, where users are asked to select the most natural face. The identity similarity is measured by $\rm{WR}_{\rm{id}}$, where users are given the ground-truth face and asked to select the most similar candidate face image.

\begin{table}[t]
\caption{\textbf{Comparison on PortraitISR.} We compare \ours~with state-of-the-art general ISR models and blending-based models. OSEDiff\_Face indicates OSEDiff trained on face images. We achieve the best performance on most of the metrics.}
\label{tab:psr}
\small
\resizebox{\textwidth}{!}{

\begin{tabular}{cccccccccccc}
\toprule
\multicolumn{1}{c}{Type} & \multicolumn{1}{c}{Method} & \multicolumn{1}{c}{PSNR$\uparrow$} & \multicolumn{1}{c}{SSIM$\uparrow$} & \multicolumn{1}{c}{LPIPS$\downarrow$} & \multicolumn{1}{c}{DISTS$\downarrow$} & \multicolumn{1}{c}{FID$\downarrow$} & \multicolumn{1}{c}{NIQE$\downarrow$} & \multicolumn{1}{c}{MUSIQ$\uparrow$} & \multicolumn{1}{c}{ID-Score$\uparrow$} & \multicolumn{1}{c}{$\rm{WR}_{\rm{id}}\uparrow$} & \multicolumn{1}{c}{$\rm{WR}_{\rm{N}}\uparrow$}  \\ 
\midrule
\multirow{2}{*}{General ISR}          
                              & OSEDiff                     &  25.19                         &     0.7802                     &       0.3287                     &    0.1755                        &                  116.80        &    4.8639                       &         66.0657                   &      0.2821      
                              &       0.21   & 0.24
                              \\
                              & PiSA-SR                     &       24.93                    &              0.7519             &             0.3506               &            0.1725                &           119.79               &                 \textbf{\color{red}4.3204  }        &         66.4237                   &          0.3118  
                              &     0.17     &  0.13
                              \\
\midrule
\multirow{2}{*}{Blending}     & RealESRGAN+Codeformer       &            24.86               &     0.7400                      &        0.4700                    &             0.2600          &   187.61  &    5.2831                     &           48.8817                &   0.1664                       
                              &      0.06    &    0.02
                              \\
                              & OSEDiff+OSEDiff\_Face       &    25.26                       &   0.7810                        &         0.3311                   &           0.1728                 &        116.86                 &          4.8791                &     64.9308                       &  0.3128                      
                              &     0.16     &  0.21
                              \\
\midrule
 \cellcolor{mygray}Portrait ISR&  \cellcolor{mygray}Ours                                   &                 \cellcolor{mygray}\textbf{\color{red}25.64}         &          \cellcolor{mygray}\textbf{\color{red}0.8060}                 &           \cellcolor{mygray}\textbf{\color{red}0.2573}                &                      \cellcolor{mygray}\textbf{\color{red}0.1398}      &             \cellcolor{mygray}\textbf{\color{red}101.13}             &            \cellcolor{mygray}4.8813               &       \cellcolor{mygray}\textbf{\color{red}67.7528}                     &                \cellcolor{mygray}\textbf{\color{red}0.3715}                   
                              &      \cellcolor{mygray}\textbf{\color{red}0.40}    &  \cellcolor{mygray}\textbf{\color{red}0.40} 
                              \\
\bottomrule

\end{tabular}
}
\end{table}

\paragraph{Implementation Details.} We initialize the base diffusion model $\Psi$ using SD 2.1 model~\citep{rombach2021highresolution}. The degradation function $\Phi$ is derived from Real-ESRGAN~\citep{wang2021real}. The face extractor is set as CVLFace~\citep{kim2024keypoint}. For the text prompt extractor, we adopt the DAPE module in~\citep{wu2024seesr}. The LoRA rank for all modules is set to 4. We train our model with AdamW~\citep{loshchilov2017decoupled} optimizer with the learning rate of $5\times10^{-5}$. The loss weighting is set as $\lambda^{\rm{fid}}=1, \lambda_{\rm{LPIPS}}^{\rm{id}}=0.8, \lambda^{\rm{id}}=4$, while the rest weights are derived from OSEDiff~\citep{wu2024one}. We train our model on eight A100 GPUs.

\begin{table}[t]
\small
\caption{\textbf{Comparison on General ISR.} We compare \ours~with state-of-the-art GISR models on the DIV2K, RealSR, and DRealSR dataset. We achieve state-of-the-art performance on some metrics while maintaining comparable on the rest metrics. We highlight the top two performances on each metric using {\color{red}red} and {\color{greyblue}blue}. `S' indicates the number of diffusion steps.}
\label{tab:isr}
\resizebox{\textwidth}{!}{

\begin{tabular}{ccccccccccc}

\toprule
\multicolumn{1}{c}{DataSet}                & Method        & \multicolumn{1}{c}{PSNR$\uparrow$} & \multicolumn{1}{c}{SSIM$\uparrow$} & \multicolumn{1}{c}{LPIPS$\downarrow$}  & \multicolumn{1}{c}{DISTS$\downarrow$} & \multicolumn{1}{c}{FID$\downarrow$} & \multicolumn{1}{c}{NIQE$\downarrow$} & \multicolumn{1}{c}{MUSIQ$\uparrow$} & \multicolumn{1}{c}{CLIPIQA$\uparrow$} & \multicolumn{1}{c}{MANIQA$\uparrow$} \\
\midrule
\multicolumn{1}{c}{\multirow{9}{*}{DIV2K}} & ResShift-S15  &   \textbf{\color{red}24.69}                    &                \textbf{\color{red}0.6175}         &          0.3374                 &                                         0.2215         &     36.01                    &            6.82              &            60.92                &      0.6089                        &                     0.5450       \\
\multicolumn{1}{c}{}                       & StableSR-S200 &     23.31                     &             0.5728              &      0.3129                     &                           0.2138                                                 &         \textbf{\color{red} 24.67}                &     4.76                      &                    65.63        &             0.6682        &0.6188         \\
\multicolumn{1}{c}{}                       & DiffBIR-S50   &       23.67                   &              0.5653            &          0.3541                 &                         0.2129                            &   30.93                      &           4.71                &          65.66                 &                0.6652             &           0.6204                 \\
\multicolumn{1}{c}{}                       & PASD-S20      &      23.14                 &         0.5489                  &         0.3607                  &                     0.2219                               &      29.32                   &               \textbf{\color{red} 4.40}          &              \textbf{\color{greyblue}68.83}             &       0.6711                     &            \textbf{\color{red}0.6484 }               \\
\multicolumn{1}{c}{}                       & SinSR-S1      &     \textbf{\color{greyblue}24.43}                      &        0.6012                  &           0.3262                &                         0.2066                           &     35.45                     &                6.02                      &             62.80                &             0.6499       & 0.5395        \\
\multicolumn{1}{c}{}                       & SeeSR-S50         &      23.71                    &            0.6045               &       0.3207                    &                       \textbf{\color{greyblue}0.1967}                             &    25.83                     &                       4.82    &               68.49            &                 \textbf{\color{greyblue}0.6857}            &               0.6239             \\
\multicolumn{1}{c}{}                       & OSEDiff-S1       &       23.72                   &       0.6108                  &      \textbf{\color{greyblue}0.2941}                     &                   0.1976                                &   26.32                      &         4.71                 &                67.97            &       0.6683                       &             0.6148               \\
\multicolumn{1}{c}{}                       & PiSA-SR-S1       &        23.87                  &         0.6058                 &          \textbf{\color{red}0.2823}                 &                      \textbf{\color{red}0.1934}                              &     \textbf{\color{greyblue}25.07 }                   &            4.55                              &              \textbf{\color{red}69.68}    &        \textbf{\color{red} 0.6927 }           &              \textbf{\color{greyblue}0.6400}              \\
\multicolumn{1}{c}{}                       & \cellcolor{mygray}Ours-S1          &          \cellcolor{mygray}23.83                &        \cellcolor{mygray}\textbf{\color{greyblue}0.6170}                  &     \cellcolor{mygray}0.3265                      &              \cellcolor{mygray}0.2102            &                      \cellcolor{mygray}28.91    &    \cellcolor{mygray}\textbf{\color{greyblue}4.45}     &    \cellcolor{mygray}64.15             &      \cellcolor{mygray}0.6445                    &                                                                                  \cellcolor{mygray}0.6162\\
\midrule
\multirow{9}{*}{RealSR}                    & ResShift-S15  &                  \textbf{\color{red}26.31}        &            \textbf{\color{greyblue}0.741}              &        0.3489                   &        0.2498                                           &                142.81          &             7.27              &                               58.10                    &     0.5450     &     0.5305                    \\
                                           & StableSR-S200 & 24.69     & 0.7052    & 0.3091      & 0.2167  & 127.20 & 5.76    & 65.42     & 0.6195             &  0.6211    \\
                                           & DiffBIR-S50   & 24.88     & 0.6673     &  0.3567      &  0.2290    &  124.56      & 5.63   & 64.66  & 0.6412   &        0.6231      \\
                                           & PASD-S20      &             25.22             &                    0.6809                                &   0.3392                       &         0.2259                  &      \textbf{\color{red}123.08  }            &    \textbf{\color{greyblue}5.18}                      &  68.74                         &     0.6502                        &   \textbf{\color{greyblue}0.6461}                         \\
                                           & SinSR-S1      &        \textbf{\color{greyblue}26.30}                  &            0.7354                                        &           0.3212               &          0.2346                 &        137.05                 &        6.31  
                                           & 60.41
                                           &        0.6204                 &          0.5389                   \\
                                           & SeeSR-S50         &         25.33                 &         0.7273                                           &             0.2985              &          0.2213                  &            125.66              &     5.38                     &    \textbf{\color{greyblue}69.37}                        &       0.6594                      &      0.6439                      \\
                                           & OSEDiff-S1       &          25.15                &         0.7341                                           &         0.2921                 &      0.2128                     &   \textbf{\color{greyblue}123.50     }                 &    5.65    
                                           & 69.09
                                           &     \textbf{\color{greyblue}0.6693}                  &       0.6339                      \\
                                           & PiSA-SR-S1       &          25.50                &         \textbf{\color{red}0.7417}                                            &            \textbf{\color{greyblue}0.2672}              &          \textbf{\color{greyblue}0.2044}                 &       124.09                  &       5.50                   &         \textbf{\color{red}70.15}               &         \textbf{\color{red}0.6702}                       &    \textbf{\color{red}0.6560}                        \\
                                           & \cellcolor{mygray}Ours-S1          &    \cellcolor{mygray}25.22                      &              \cellcolor{mygray}0.7238                                      &            \cellcolor{mygray}\textbf{\color{red}0.2671}              &         \cellcolor{mygray}\textbf{\color{red}0.1943}   &          \cellcolor{mygray}131.07     &     \cellcolor{mygray}\textbf{\color{red}4.86}                    &  \cellcolor{mygray} 65.05                       &                 \cellcolor{mygray}0.6296                                      &     \cellcolor{mygray}0.6332                       \\
\midrule
\multirow{9}{*}{DRealSR}                   & ResShift-S15  &    \textbf{\color{red}28.45}                      &     0.7632                     &    0.4073                      &                0.2700          &         175.92                   &               8.28      &            49.86        &            0.5259                                                       &    0.4573                        \\
                                           & StableSR-S200 &             28.04             &           0.7460     &                   0.3354                 &         0.2287                  &        147.03                    &        6.51   
                                           &       58.50
                                           &       0.6171                  &   0.5602                         \\
                                           & DiffBIR-S50   &             26.84             &              0.6660                                      &
                                           0.4446
                                           &
                                           0.2706 &                       167.38   &                     6.02     &   
                                            60.68&
                                           0.6292       &                 0.5902                                       \\
                                           & PASD-S20      &             27.48             &             0.7051             &         0.3854                  &    0.2535                                                &         157.36                &      \textbf{\color{red}5.57    }                &       64.55                    &          0.6714                   &        \textbf{\color{greyblue}0.6130}                    \\
                                           & SinSR-S1      &         \textbf{\color{greyblue}28.41}                 &           0.7495               &          0.3741                 &                            0.2488                       &             177.05            &     7.02                     &            55.34               &          0.6367                   &                0.4898             \\
                                           & SeeSR-S50         &         28.26                 &        0.7698                  &      0.3197                     &        0.2306                                            &          149.86               &        6.52                  &       64.84                    &       0.6672                      &   0.6026                        \\
                                           & OSEDiff-S1       &                         27.92 &     \textbf{\color{red}0.7835}                    &                    0.2968       &                              \textbf{\color{greyblue}0.2165}                      &               \textbf{\color{greyblue}135.29}          &                  6.49        &     \textbf{\color{greyblue} 64.65}                  &                     \textbf{\color{greyblue}0.6963}           &            0.5899                \\
                                           & PiSA-SR-S1       &    28.31                      &           0.7804               &                    \textbf{\color{greyblue}0.2960}       &                             0.2169                       &                  \textbf{\color{red} 130.61}      &      6.20                    &     \textbf{\color{red}66.11} &  \textbf{\color{red}0.6970}                               &    \textbf{\color{red} 0.6156 }                     \\
                                           & \cellcolor{mygray}Ours-S1          &   \cellcolor{mygray}27.70                       &     \cellcolor{mygray}\textbf{\color{greyblue}0.7826}                     &  \cellcolor{mygray}\textbf{\color{red}0.2885}                         &             \cellcolor{mygray}\textbf{\color{red}0.2038}                                       &                        \cellcolor{mygray}137.48
                                           &        \cellcolor{mygray}\textbf{\color{greyblue}5.81 }                 &                 \cellcolor{mygray}61.59          &         \cellcolor{mygray}0.6609                    &            \cellcolor{mygray}0.5970        \\       
\bottomrule
\end{tabular}
}
\vspace{-3mm}
\end{table}

\subsection{Comparison on PortraitISR}

We compare our model with the baselines on \ourdata-Te dataset. The quantitative results are shown in Tab.~\ref{tab:psr}. We observe that \ours~continuously achieves state-of-the-art performance on most of the metrics. The qualitative results are illustrated in Fig~\ref{fig:compare}, from which we observe that (1) the general ISR approaches can achieve good enhancement in non-facial regions, but they usually suffer from hallucinated faces. (2) The hybrid blending approaches, in contrast, can preserve more detail and identity of the face, but can easily cause inconsistency around the boundary regions.

\subsection{Comparison on GISR and FISR Tasks}

In this section, we evaluate \ours~on the general image super-resolution (GISR) and face image super-resolution (FISR) datasets. The experimental results show that our method can generalize to other datasets and achieves competitive performance on all the tasks.

\begin{table}[t]
\caption{\textbf{Ablation Studies.} We ablate the key component of \ours. Compare the first and fourth row, adding the face-aware loss significantly improves both the photometric criteria of the entire image and the identity similarity of the face region. Adding $L_F^{\rm fid}$ alone (second row) slightly improves the identity similarity. Using $L_F^{\rm id}$ alone (third row) can significantly improve the identity similarity, at the cost of introducing significant blur effects, shown by the low non-reference metrics and \cref{fig:ablation}. Comparing the fourth and fifth row, adding a reference image can slightly improve identity similarity.}
\label{tab:abl}
\small
\resizebox{\textwidth}{!}{

\begin{tabular}{ccccccccccc}
\toprule
$\mL_\mF^{\rm{fid}}$ & $\mL_\mF^{\rm{id}}$ & w/ $\Mat x_r$ & PSNR $\uparrow$    & SSIM $\uparrow$  & LPIPS $\downarrow$ & DISTS$\downarrow$  & FID$\downarrow$ & NIQE$\downarrow$   & MUSIQ $\uparrow$  & ID-Score$\uparrow$ \\
\midrule
   &    &     & 25.07 & 0.8059 & 0.2755 & 0.1533 &   111.16  & 4.9286 & 69.2658 & 0.2600   \\
\checkmark  &    &     & 25.65 & 0.8113 & 0.2770 & 0.1631 & 110.44    & 5.1377 & 67.4218 & 0.3010   \\
   & \checkmark  &     & 25.85 & 0.8122 & 0.2680 & 0.1527 &  104.82   & 5.7209 & 64.5272 &  0.4348        \\
\checkmark & \checkmark  &             &   25.74     &  0.8040        &    0.2517    &   0.1360& 99.55 &    4.5785    &      69.0616   &      0.3634    \\
\checkmark  & \checkmark  & \checkmark   &  25.64        &          0.8060             &           0.2573                &                      0.1398      &            101.13             &            4.8813               &       67.7528                     &                0.3715           \\
\bottomrule
\end{tabular}
}
\vspace{-4mm}
\end{table}

\paragraph{Comparison on GISR task.} We show that our model can generalize to the GISR task. We compare our model with state-of-the-art GISR methods on real-world image super-resolution datasets. The results are shown in~\cref{tab:isr}. We achieve state-of-the-art in multiple metrics (NIQE on RealSR, LPIPS on DRealSR, etc.). Regarding the rest of the metric, we reach comparable performance to the state-of-the-art general image super-resolution approaches.
\paragraph{Comparison on FISR task.} 
\begin{wrapfigure}{r}{0.35\textwidth}
    \centering
    \vspace{-3mm}
    \includegraphics[width=0.3\textwidth]{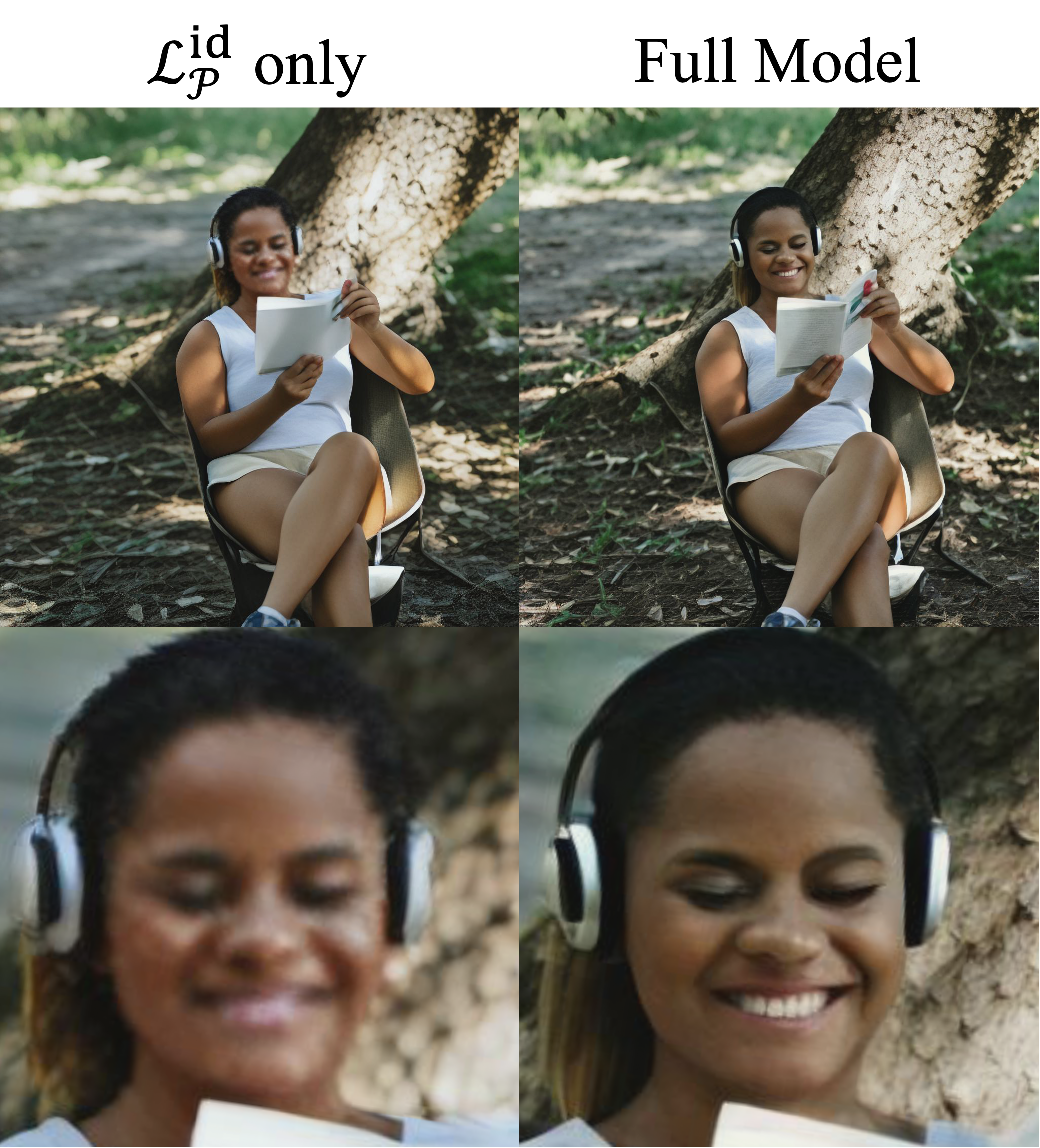}\vspace{-3mm}
    \caption{\textbf{Ablation Studies.} Using the identity loss only, though, improves the identity-preservation ability, but will lead to over-smooth and blurred faces.}
    \vspace{-9mm}
    \label{fig:ablation}
\end{wrapfigure}

We also show the \ours's ability to generalize to the face image super-resolution task.
We employed the widely used synthetic face dataset CelebA-Test~\citep{liu2015faceattributes} as the testing set. CelebA-Test consists of 3000 cropped and aligned faces.
Following the settings in~\citep{tsai2024dual}, we evaluated \ours on the aligned face dataset, and compared it with the state-of-the-art specialist FISR models. The results can be found in ~\cref{tab:fsr}. We achieve the best performance on most metrics and comparable performance on the rest of the metrics.

\subsection{Ablation Studies}

\vspace{-2mm}
\paragraph{Ablation on Face Aware Losses.} We ablate each component of the face-aware losses. From Tab.~\ref{tab:abl} we observe that without face-aware supervision (first row ) will lead to better no-reference metrics, while the identity of the face is not preserved. Adding $\gL_\gP^{\rm{fid}}$ (second row )can improve the ID-Score. Using the identity loss alone (third row) can encourage the model to keep the identity, but the image quality will significantly drop. The reason is that the identity loss employed a feature similarity score, which may encourage the model to produce blurred and smoothed faces, as shown in Fig.~\ref{fig:ablation}.
\paragraph{Ablation on Reference Images.} Our model can restore portrait images with or without a reference image. We compare the performance of the models with reference (last row) and without reference (fourth row). We observe that the reference image slightly improves the identity score but reduces image quality by a small margin.

\section{Related Works}
\subsection{General Image Super-Resolution}

General image super-resolution (GISR) aims to restore HQ images from LQ inputs. SRCNN~\citep{dong2015image} first employed convolutional neural networks to solve the GISR problem. Following approaches further developed more powerful deep learning approaches for GISR by improving CNN structure~\citep{lim2017enhanced, zhang2018residual,zhang2018image}, introducing attentions~\citep{dai2019second, zhang2022efficient} or transformers~\citep{chen2023activating,chen2023dual,liang2021swinir,chen2021pre}, or leveraging multi-scale features~\citep{li2018multi, gao2019multi}. With the development of generative models, SR-by-generating became an important branch of GISR. SRGan~\citep{ledig2017photo} first introduced the generative prior of GAN~\citep{goodfellow2020generative} into image super-resolution. The generative prior enables SR models to produce better texture details. The following works~\citep{zhang2021designing, wang2021real, xie2023desra, liang2022efficient, liang2022details, zhang2021designing} push super-resolution to real-world images, whose degradation is unknown to the algorithm, by developing complex degradation pipelines. Recently, many works have explored leveraging generative priors in diffusion models for GISR~\citep{wu2024seesr, yue2023resshift, yang2024pixel, wu2024one, wang2024sinsr, sun2024pixel, lin2024diffbir}. Although achieving great success regarding the general image quality, just like other GISR methods, they often produce unnatural and low-fidelity faces for portrait images. 

\subsection{Face Image Super-Resolution}

Face image super-resolution (FISR) is a task that specifically focuses on aligned facial images. Compared to general natural images, humans are more sensitive to the details of faces. Thus, FISR requires more fine-grained restoration of details.  Many designs have been made to achieve high-fidelity face image super-resolution. For example, some approaches~\citep{chen2018fsrnet, shen2018deep, kim2019progressive, yu2018super, zhang2022multi} employed face structure or landmarks to provide a geometric prior in FISR. Reference-based methods~\citep{zhang2024instantrestore, li2020blind, li2020enhanced, li2018learning, chong2025copy} leverage one or more reference images to alleviate ambiguity and preserve more identity while conducting super-resolution. Quantized codebook~\citep{wang2024osdface, zhou2022towards, tsai2024dual} is an effective way to model general face features. By querying the codebook, these kinds of methods can generate high-quality face images from the low-quality ones. Similar to GISR, there are also a large number of FISR models that leverage the generative prior from GAN~\citep{yang2020hifacegan, wan2020bringing, menon2020pulse, wang2021towards} or diffusion models~\citep{wang2024osdface, yang2023pgdiff}. Most FISR methods rely on the prior where the face images are roughly aligned with a template face. When applying to a portrait image, another GISR model is required to enhance the non-facial regions. 

\subsection{Diffusion Models in Image Super-Resolution}

Recent success on pre-trained large diffusion model~\citep{rombach2021highresolution, zhang2023adding, esser2024scaling, peebles2023scalable} has significantly facilitated a large number of visual tasks~\citep{li20244k4dgen, lugmayr2022repaint, saxena2023surprising, baranchuk2021label}. StableSR~\citep{wang2024exploiting} employs a trainable adapter to leverage the generative prior of pre-trained diffusion models. SeD~\citep{li2024sed} combined GANs and diffusion models to produce more photo-realistic images. PASD~\citep{yang2024pixel} leverages both high- and low-level features to enable diffusion models to perceive image
local structures at a pixel-wise level. Further, several efforts have been put into reducing the diffusion steps~\citep{wu2024one, wang2024sinsr,sun2024pixel}. Further, PiSA-SR~\citep{sun2024pixel} provides a flexible trade-off between the pixel-wise fidelity and semantic-level details by 
introducing two adjustable guidance scales on two
LoRA modules.

\section{Conclusion}
\label{sec:conc}
 We introduce the task of PortraitISR, which aims to enhance portrait images that consist of both a human face and other components, such as the human body and natural background. Existing ISR approaches either general images or aligned face images, which usually suffer from low-fidelity face restoration or inconsistency around boundaries when applied to portrait images. We propose \ours, the first end-to-end PortraitISR framework. We designed a face-aware region loss and a reference-guidance structure to improve facial restoration quality. We further build \ourdata, a high-resolution portrait data, facilitating future research and benchmarking in PortraitISR tasks. Experimental results on multiple datasets show that we achieve state-of-the-art performance on the PortraitISR task and competitive performance on the general ISR and face ISR tasks. Our proposed \ourdata~provides high-quality portrait data, which can potentially facilitate various future research and benchmarking in fields like portrait super-resolution, generation, matting, among others.

\bibliography{main}
\bibliographystyle{iclr2026_conference}

\appendix
\newpage
\section{Appendix Overview}
Due to space constraints in the main draft, we include implementation details, the data curation, and experimental results in the appendix. Specifically, in Sec. \ref{sec:sup-imp}, we offer further explanation of the implementation of our framework and the experiments. In Sec.~\ref{sec:sup-data}, we present the details of \ourdata. Finally, in Sec. \ref{sec:sup-res}, we present additional visual results of the main experiments.

\section{Implementation Details}
\label{sec:sup-imp}
\paragraph{Backbone Settings.} We use the Stable Diffusion 2.1 model~\citep{rombach2021highresolution} as the base model for LoRA finetuning. We employ the CVLFace~\citep{kim2024keypoint} as the face feature extractor. The DAPE model as in~\citep{wu2024seesr} is utilized to extract the language description of the input LQ images, while the negative prompt is fixed as ``painting, oil painting, illustration, drawing, art, sketch, cartoon, CG Style, 3D render, unreal engine, blurring, dirty, messy, worst quality, low quality, frames, watermark, signature, jpeg artifacts, deformed, lowres, over-smooth'' to avoid synthetic and low-quality styles. 

\begin{figure}[h]
    \centering
    \includegraphics[width=\textwidth]{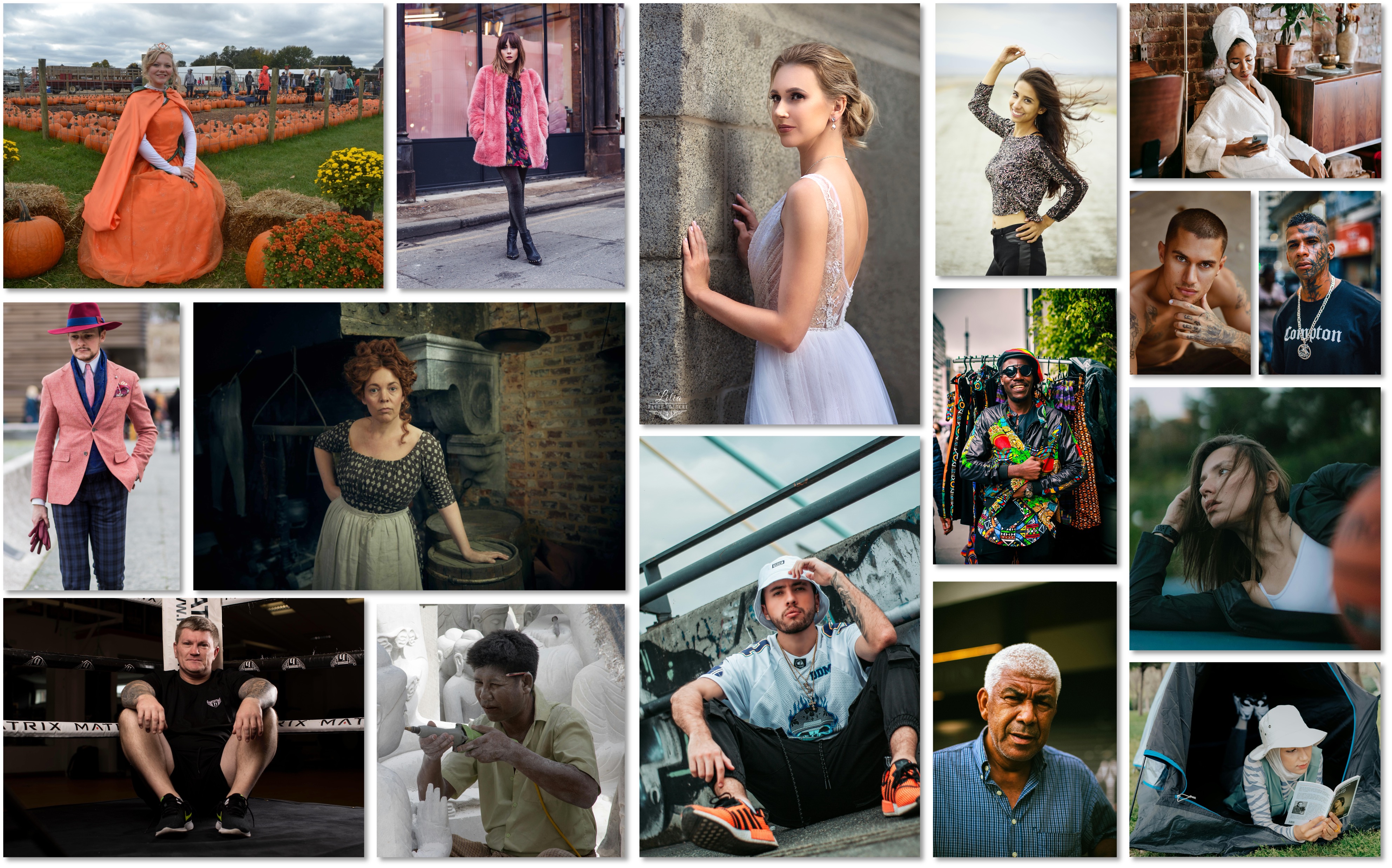}\vspace{-1mm}
    \caption{Visualization of some portrait images sampled from \ourdata.}
    \label{fig:supp-data-vis}
\end{figure}
\paragraph{Model details.} We concatenated the LQ latent (4 channels), the reference latent (4 channels), and the resize binary mask (one channel) before sending them to the denoise UNet. We modify the `conv-in' layer to enable the denoising UNet with a nine-channel input. The first four channels for the new convolution filter are initialized from the base model, while the five channels are initialized to zero. The bias of the convolution layer is initialized from the base model.
\paragraph{Face Alignment.} We leverage FaceLib to detect face landmarks following~\citep{zhou2022towards}. We then estimate the affine transformation between the detected landmark and a template landmark using the `estimateAffinePartial2D' in OpenCV. The wrapping transformation is implemented using `affine\_grid' and `grid\_sample' in PyTorch to ensure the procedure is differentiable.
\paragraph{Training and Evaluation.} The training and inference resolution is at $1024\times 1024$. We train the model for 150k steps in two stages. The first consists of 40k steps, the ratio of training data is $\rm{PortraitSR-4K}:\rm{ffhq}:\rm{lsdir}:\rm{div2k}=0.15:0.05:1.7:0.3$. The second stage consists of the remaining 110k steps, and the ratio of mixed training data is $\rm{PortraitSR-4K}:\rm{ffhq}:\rm{lsdir}:\rm{div2k}=1.5:0.5:1.7:0.3$. The probability of dropping the reference image in both stages is 0.2. For evaluation metrics, we use  pyiqa~\citep{pyiqa} to calculate the PSNR, SSIM, LPIPS, DISTS, NIQE, MUSIQ, CLIPIQA, and MANIQA-pipal scores. The rest of the metrics are calculated using their original codes.
\paragraph{User Study Details.}
We collected 36 questionnaires for the win rate on similarity ($\rm{WR}_{\rm{id}}$), each questionnaire contains 58 questions randomly sampled from the testset. For each question, participants are given five face images produced by the five approaches in Tab.~\ref{tab:psr}, and one ground-truth face image. Participants are asked to select the image that is most similar to the ground-truth face image. We collected 28 questionnaires for the win rate on naturalness ($\rm{WR}_{\rm{N}}$), each questionnaire contains 58 questions randomly sampled from the testset. For each question, participants are provided with five face images produced by the five models. The ground-truth image is not provided. Participants are asked to select the most natural face out of the five given images. In both questionnaires, the order of options is randomly shuffled.

\section{\ourdata~Details}
\label{sec:sup-data}
We selected the images that are at least $3840 \times 2160$, with the longest side exceeding 3840 pixels, as candidates. The aspect ratio of the images ranges from 0.6 to 1.6. We further filter the images using the Q-align~\citep{wu2023q} aesthetic and quality scores. 
We detect the face in each image using FaceLib. We drop faces whose distance between the two eyes is less than 64 pixels. We leverage ~CVLFace\citep{kim2024keypoint} to estimate the similarity of face pairs. We exclude face pairs whose similarity is below 0.65. The reference face is cropped and aligned using the face alignment techniques described in Sec.~\ref{sec:sup-imp}. We visualize several portrait images  from \ourdata~in Fig.~\ref{fig:supp-data-vis}.

\section{Experimental Results}
\label{sec:sup-res}
\begin{figure}
    \centering
    \includegraphics[width=\textwidth]{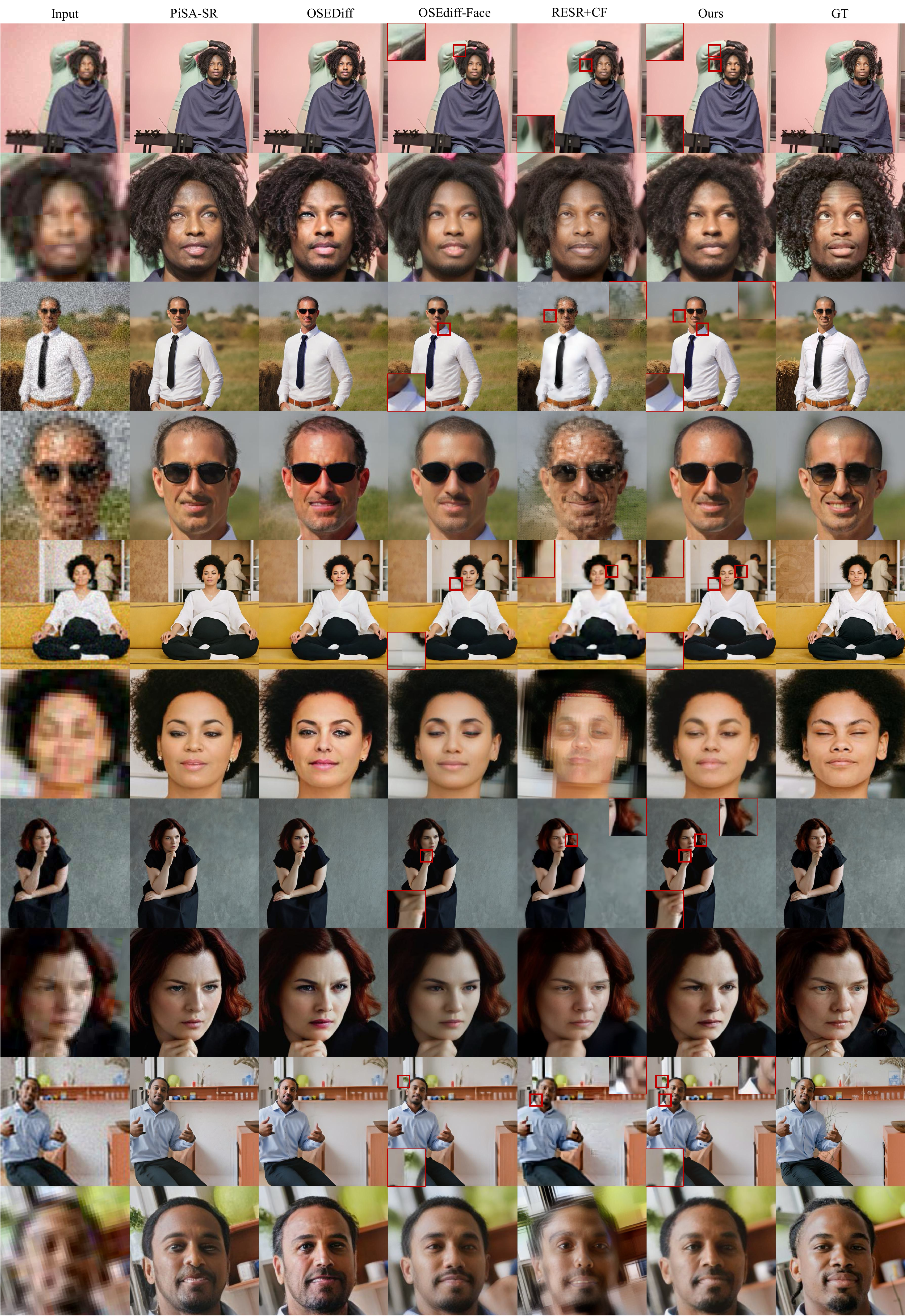}\vspace{-1mm}
    \caption{\textbf{Visual Comparison on PortraitISR.} OSEDiff-Face is a blending-based approach, the background is handled by OSEDiff~\citep{wu2024one} trained for general image super-resolution, and the face region is processed by a specialist DSEDiff model that is trained on the face dataset. Similarly, RESR+CF is the blending-based approach which combines Real-ESRGAN~\citep{wang2021real} and CodeFormer~\citep{zhou2022towards}.}
    \label{fig:supp-comp-pisr}
\end{figure}

\begin{figure}
    \centering
\includegraphics[width=\textwidth]{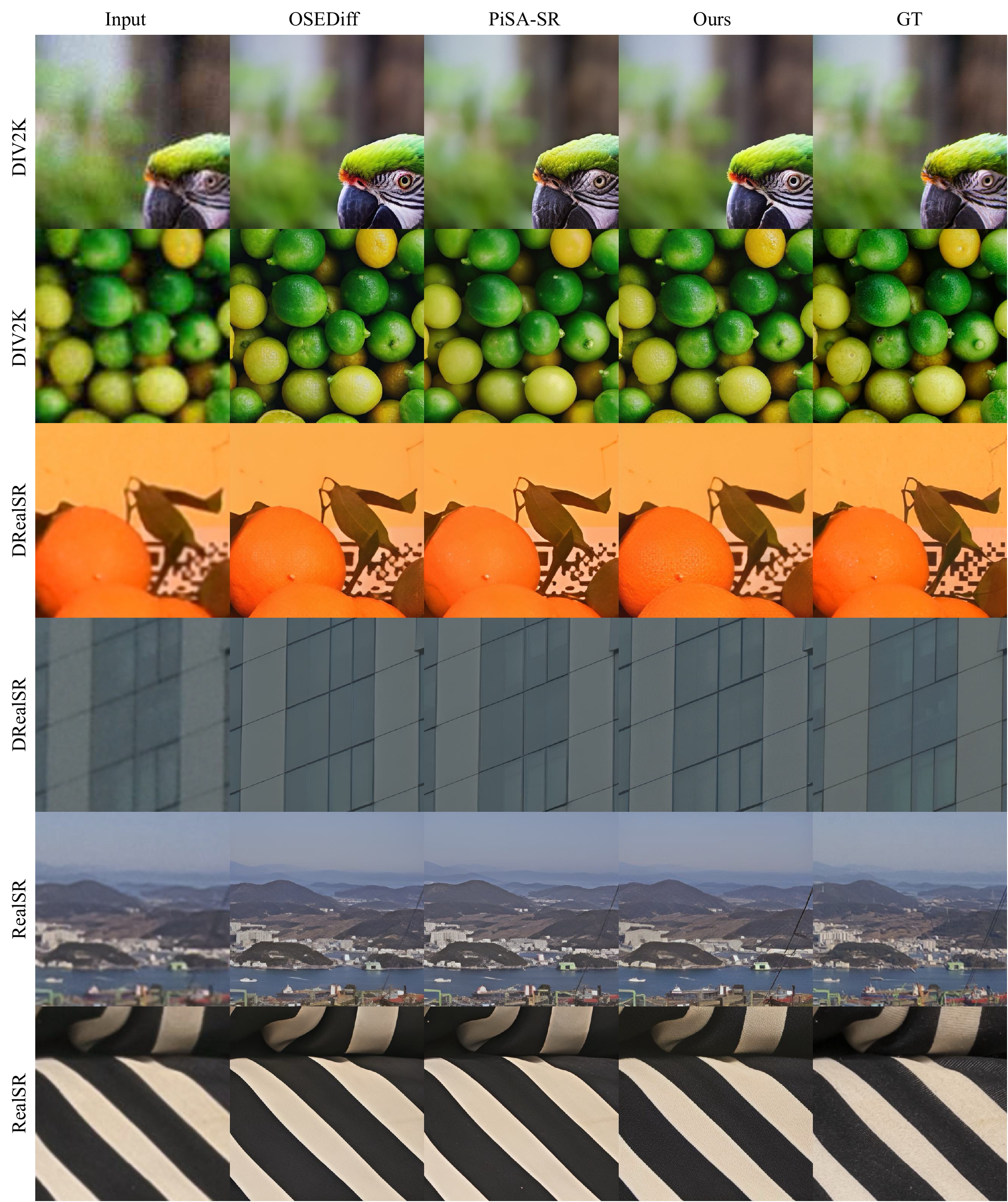}\vspace{-1mm}
    \caption{\textbf{Visual Comparison on GISR.} Our model achieves competitive results.}
    \label{fig:supp-comp-gisr}
\end{figure}

\begin{figure}
    \centering
\includegraphics[width=\textwidth]{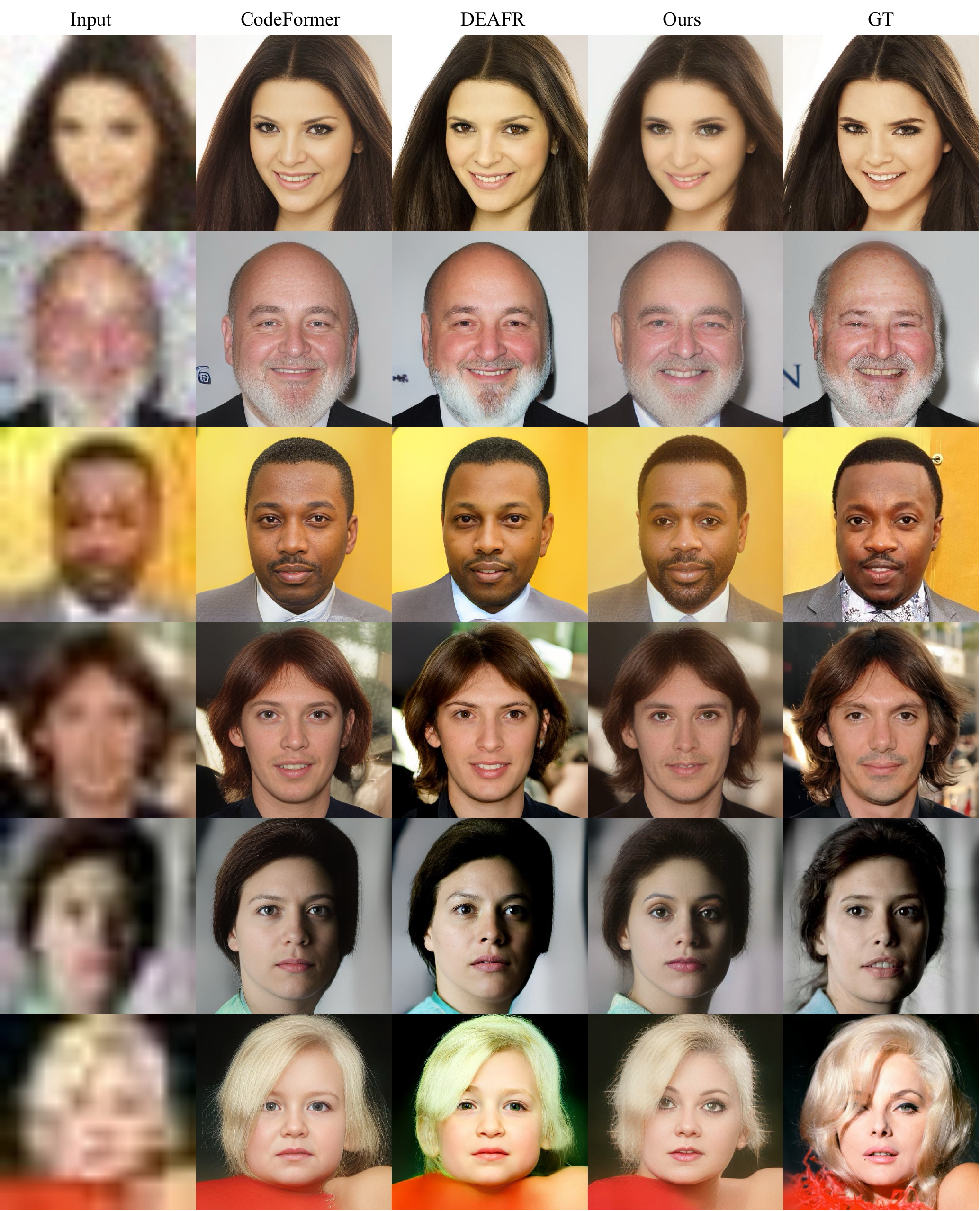}\vspace{-1mm}
    \caption{\textbf{Visual Comparison on FISR.}  Our model achieves performance comparable to that of the FISR specialists.}
    \label{fig:supp-comp-fisr}
\end{figure}

We show more visualization results on the PortraitISR task in Fig.~\ref{fig:supp-comp-pisr}. As shown in the figure, our model achieves better face fidelity compared to general ISR models, while avoiding inconsistent borders compared to blending-based approaches.  The visual results for the GISR task are shown in Fig.~\ref{fig:supp-comp-gisr}, which demonstrate our competitive performance compared to the state-of-the-art models. The visual results for the FISR task are shown in Fig.~\ref{fig:supp-comp-fisr}. Our model achieves performance comparable to that of the specialists for face image super-resolution.

\end{document}